\documentclass{article}

\usepackage[preprint]{neurips_2022}

\usepackage[utf8]{inputenc} 
\usepackage[T1]{fontenc}    
\usepackage{hyperref}       
\usepackage{url}            
\usepackage{booktabs}       
\usepackage{amsfonts}       
\usepackage{nicefrac}       
\usepackage{microtype}      
\usepackage{xcolor}         
\usepackage{amsmath}
\usepackage{bm}
\usepackage{graphicx}
\usepackage{epstopdf}
\usepackage{algorithm}
\usepackage{algpseudocode}
\usepackage{subfigure}
\usepackage{multirow}

\title{Challenges in Binary Classification}

\author{%
  Pengbo Yang \\
  Beijing Key Lab of Traffic Data Analysis and Mining\\
  Beijing Jiaotong University\\
  \texttt{18112035@bjtu.edu.cn} \\
  \And
  Jian Yu\\
  Beijing Key Lab of Traffic Data Analysis and Mining\\
  Beijing Jiaotong University\\
  \texttt{jianyu@bjtu.edu.cn} \\
}

\begin{document}

\maketitle

\begin{abstract}
 Binary Classification plays an important role in machine learning. For linear classification, SVM is the optimal binary classification method. For nonlinear classification, the SVM algorithm needs to complete the classification task by using the kernel function. Although the SVM algorithm with kernel function is very effective, the selection of kernel function is empirical, which means that the kernel function may not be optimal. Therefore, it is worth studying how to obtain an optimal binary classifier.
  
 In this paper, the problem of finding the optimal binary classifier is considered as a variational problem.  We design the objective function of this variational problem through the max-min problem of the (Euclidean) distance between two classes. For linear classification, it can be deduced that SVM is a special case of this variational problem framework. For Euclidean distance, it is proved that the proposed variational problem has some limitations for nonlinear classification. Therefore, how to design a more appropriate objective function to find the optimal binary classifier is still an open problem. Further, it's discussed some challenges and problems in finding the optimal classifier.
\end{abstract}

\section{Introduction}
For supervised learning, binary classification plays a key role as multi-class classification can be turned into binary classification \citep{mathur2008multiclass}.  Therefore, it is very important to deal with binary classification.  

Support Vector Machine (SVM) stands out as one of the most prominent algorithms for linear classification tasks \citep{cortes1995support}. SVM works by finding the optimal hyperplane that separates the classes in the feature space. Its effectiveness lies in its ability to maximize the margin between classes, making it robust to noise and outliers. However, the assumption of linear separability may not hold true for many real-world datasets. In fact, the probability of a binary classification problem being linearly separable is often quite low. This limitation necessitates the exploration of nonlinear classification approaches. Nonlinear classification can be achieved within the SVM framework through the use of kernel functions \citep{amari1999improving}. Despite the effectiveness of SVMs with kernel functions, selecting the optimal kernel remains a challenging and unsolved problem. 


This paper introduces a variational problem based on the Euclidean distance between two classes to find the optimal binary classifier. Within the framework of this variational problem, the objective function of the support vector machine can be derived for linear classification. Further, we study whether the variational framework proposed in this paper can find the optimal classifier for nonlinear classification problems. Unfortunately, this variational problem cannot find the optimal classification function under the condition that the quadratic surface is used as the classifier.

Finally, we explore the challenges and issues of designing a better objective function to find the optimal binary classifier. We design a heuristic metric objective based on the integral of the density function to find the optimal binary classifier. However, the existence and uniqueness of the solution to this objective function still need to be proven. Secondly, how to optimize and solve this objective function is also a thorny problem.

\section{Variation Problem For Finding The Optimal Binary Classifier}
\label{2}
\textbf{Problem Statement.} Suppose that $N$ examples $x_k$ with labels $y_k$, where $\forall k,x_k\in R^p$, $y_k\in\{-1,1\}$.  Strictly speaking, when $x_k\in$ class $\Omega _{+}$, $y_k=1$; when $x_k\in$ class $\Omega _{-}$, $y_k=-1$.  It is a reasonable assumption that exists a decision function $f(x)=0$, such that $\forall k, f(x_k)y_k\geq 1$, where $f\in C^{\infty}(R^p)$ is an infinitely differentiable function: $R^p\mapsto R$.  The above data settings are the basic assumptions of the support vector machine model. Let $\Omega _+^f$ and $\Omega _-^f$ represent the support vector sets of positive and negative samples under function $f$, i.e., $\Omega _+^f=\{x \in R^p |f(x)=1\}$, and $\Omega _-^f=\{x^{'}\in R^p|f(x^{'})=-1\}$. In order to find the optimal binary classifier $f$, it is necessary to design a metric $d(\Omega _-^f,\Omega _+^f)$, which reflects the distance between $\Omega _-^f$ and $\Omega _+^f$.



\textbf{Open Problem.} \quad How to design an appropriate $d(\Omega _-^f,\Omega _+^f)$ to find the optimal classifier $f$?

For simplicity, set $d(\Omega _-^f,\Omega _+^f)$ to the minimum Euclidean distance between $\Omega _-^f$ and $\Omega _+^f$. It is easy to see that the optimal function $f(x)$ should satisfy the Formula (\ref{vsvmobj}), which guarantees that the decision function has the maximal margin between positive and negative classes. 

\begin{equation}\label{vsvmobj}
f^* = \max_{f\in C^{\infty}(R^p)} d(\Omega _-^f,\Omega _+^f)=\max_{f\in C^{\infty}(R^p)} \min_{f(x)=1,f(x^{'})=-1} \|x-x^{'}\|^2, \forall k, f(x_k)y_k\geq 1.
\end{equation}
where $x$ and $x^{'}$ are support vectors from different classes. The max-min problem in Formula (\ref{vsvmobj}) is a variational problem.

Next, we need to explain whether the optimal binary classifier can be obtained by optimizing the variational problem proposed above. First, it's verified that the  objective function defined in Equation (\ref{vsvmobj}) can derive the results of SVM \citep{cortes1995support} when the data is linearly separable. Secondly, we discussed whether the above variational problem can find an optimal solution when using a quadratic function as a classifier.



\subsection{The optimal linear binary classifier: SVM}
If the data is linearly separable, the search space of $f$ is reduced from the set of $C^{\infty}$ functions to the set of linear functions in $R^P$ space. Then we consider $f(x)=w^Tx+b$. Using the Lagrange's multiplier method to solve the variation problem of Equation (\ref{vsvmobj}), the final objective function is:

\begin{equation}\label{vsvmobjlin}
f^*=max_{w,b}\frac{4}{\|w\|^2}, \forall k, f(x_k)y_k\geq 1.
\end{equation}
Please refer to Appendix \ref{appendixA} for the specific derivation process. The objective function defined by Equation (\ref{vsvmobjlin}) is the objective function of SVM \citep{cortes1995support}. Therefore, SVM is a special case of the variational problem defined in Equation (\ref{vsvmobj}).

\subsection{Quadratic surfaces as classification boundaries}
Here, we limit the search space of the classification function $f$ to all quadratic functions in $R^P$ space. Assume that the classification boundary equation is $f(x)=x^TAx+b^Tx+C=0$, where $x\in R^p$, $b\in R^p$, and $C$ is a constant. We focus on $A$ because $A$ determines the shape of the quadric surface. Set $A$ as a $p$-order square matrix that can be diagonalized, that is, $P^TAP=\Lambda$, where $\Lambda$ is a diagonal matrix with the diagonal elements $l_1\ge l_2\ge ...\ge l_p$, and $P^TP=1$. Let $x=P\tilde{x}$, then the quadratic surface square $f(x)$ can be converted into a standard surface equation $f(\tilde{x})$:

\begin{eqnarray}
\begin{aligned}
f(x)=x^TAx+b^Tx+C & = \tilde{x}^TP^TAP\tilde{x}+b^TP\tilde{x}+C \\ 
& = \tilde{x}^T \Lambda \tilde{x}+\tilde{b}^T\tilde{x}+C \\
&= f(\tilde{x})
\end{aligned}
\end{eqnarray}
where, $\tilde{b}=P^{T}b$. Next, we consider several different shapes of quadratic surfaces by setting $\Lambda$ differently:

(1) \textbf{Ellipsoid.} Assume that all eigenvalues of $A$ are greater than 0, that is, $l_1\ge l_2\ge ...\ge l_{p} >0$. According to the variation problem we defined, the objective function we need to optimize at this time is as follows:

\begin{equation}
\label{e_loss}
f^*=\max_{f(x)=x^TAx+b^Tx+C} (\alpha -\beta )^2l_1^{-1}, \quad   \forall k, f(x_k)y_k\geq 1
\end{equation}
where $\alpha =\sqrt{\frac{1}{4} \tilde{b}^T\Lambda^{-1}\tilde{b}-C+1}$ and $\beta =\sqrt{\frac{1}{4} \tilde{b}^T\Lambda^{-1}\tilde{b}-C-1}$. The physical meaning of the maximized objective in the above formula is the distance along the shortest axis of the ellipsoid. For the specific proof process, please refer to the Appendix \ref{ellipsoid surface}. The parameters $A, b$, and $C$ of $f(x)$ can be further solved using the Lagrange multiplier method and KKT conditions.

Something to note here is that if $l_1\ge l_2\ge ...l_d > 0$, $l_{d+1}=...=l_{p} =0$, and $\tilde{x}_{[d+1:p]}=0$, then such a quadratic surface will be a degenerate ellipsoid $f(\tilde{x}) = \tilde{x}_{[1:d]}^T \Lambda_{[1:d]} \tilde{x}_{[1:d]}+\tilde{b}_{[1:d]}^T\tilde{x}_{[1:d]}+C$, where $\tilde{x}_{[1:d]}=[\tilde{x}_1, \tilde{x}_2,..., \tilde{x}_d]^T$, and $\Lambda_{[1:d]}=diag([l_1, l_2, ...,l_d])$. The optimal classifier can also be solved using an optimization objective of Equation (\ref{e_loss}).

(2) \textbf{Paraboloid.} If $l_1\ge l_2\ge ...l_d > 0$, $l_{d+1}=...=l_{p} =0$, and $\tilde{x}_{[d+1:p]}\ne0$, then such a quadratic surface will be a parabolic quadratic surface. Then the classification boundary equation becomes $f(\tilde{x}) = \tilde{x}_{[1:d]}^T \Lambda_{[1:d]} \tilde{x}_{[1:d]}+\tilde{b}^T\tilde{x}+C =0$. However, we found the following results:

\begin{equation}\label{v}
d(\Omega _-^f,\Omega _+^f)= \min_{f(x)=1,f(x^{'})=-1} \|x-x^{'}\|^2=0
\end{equation}
For the specific proof process, please refer to Appendix \ref{paraboloid surface}. The result of Equation (\ref{v}) shows that the variational problem defined in Equation (\ref{vsvmobj}) cannot find the optimal solution to the binary classification problem when the parabolic surface is used as the classification boundary.  Therefore, the variation problem defined based on Euclidean distance may have certain flaws.


(3) \textbf{Hyperboloid.} If the eigenvalues of matrix $A$ are $l_1\ge l_2\ge ...l_d > 0 > l_{d+1} \ge ... \ge l_{p}$, then the classification boundary at this time is a hyperbolic surface. We proved in Appendix \ref{hyperbolic surface} that for this type of nonlinear binary classification problem, $d(\Omega _-^f,\Omega _+^f)=0$. This shows that the variational problem defined in Equation (\ref{vsvmobj}) cannot solve the hyperbolic classifier function either.

\section{Challenges in Binary Classification}
Through the analysis in Section \ref{2}, we know that the variational problem defined based on Euclidean distance cannot be well generalized to nonlinear classification functions. It may be because the objective function in Equation (\ref{vsvmobj}) only considers the distance between the two nearest points, which is a local measure. We can consider the area integral $\int_{-1\le f(x)\le 1}d(x)$ between $f(x)=-1$ and $f(x)=1$ as a new metric to replace the Euclidean distance metric. However, the area integral $\int_{-1\le f(x)\le 1}d(x)$ is likely to be infinite, so it will not be possible to obtain the optimal classifier $f$ by optimizing this metric. Therefore, it is necessary to introduce an integrand here to make the area integral become finite.

We know that the integral of the Gaussian distribution $G(x)$ is $\int_{R^p} G(x)d(x)=1$. Then the integral $\int_{-1\le f(x)\le 1} G(x)d(x)$ of the Gaussian distribution in the region between $f(x)=-1$ and $f(x)=1$ is a finite value. However, $G(x)$ is a function that is independent of data. Therefore, we consider introducing data distribution $\rho(x)$. Based on the data distribution $\rho(x)$, the following objectives function can be designed:
\begin{equation}\label{area integral}
f^* = \max_{f\in C^{\infty}(R^p)} d(\Omega _-^f,\Omega _+^f) = \max_{f\in C^{\infty}(R^p)} \int_{-1\le f(x)\le 1}\rho (x)dx, \forall k, f(x_k)y_k\geq 1.
\end{equation}
where $\int_{R^p} \rho (x)dx=1$, and $\forall k, \rho(x_k) > 0$. At this time, there are two unknown variables $\rho(x)$ and $f(x)$ in Equation (\ref{area integral}), so it is very difficult to optimize. The existence and uniqueness of the solution to Equation (\ref{area integral}) need further verification. However, once we solve the above equation, then we will have both a generative model and a classification model.

\section{Conclusion}
In this paper, we propose a variational problem based on Euclidean distance to solve the optimal binary classifier. For linear separability, the variation problem can deduce the objective function of SVM. Additionally, it is demonstrated that the proposed variational problem based on Euclidean distance has some limitations for nonlinear classification scenarios. Therefore, a better objective function needs to be designed to complete the binary classification task, which is still an open problem. Finally, we conjecture that when introducing data distribution, the variational problem based on area metric can be used to find the best binary classifier.

\bibliographystyle{unsrtnat}
\bibliography{paper}

\begin{thebibliography}{3}
\providecommand{\natexlab}[1]{#1}
\providecommand{\url}[1]{\texttt{#1}}
\expandafter\ifx\csname urlstyle\endcsname\relax
  \providecommand{\doi}[1]{doi: #1}\else
  \providecommand{\doi}{doi: \begingroup \urlstyle{rm}\Url}\fi

\bibitem[Mathur and Foody(2008)]{mathur2008multiclass}
Ajay Mathur and Giles~M Foody.
\newblock Multiclass and binary svm classification: Implications for training and classification users.
\newblock \emph{IEEE Geoscience and remote sensing letters}, 5\penalty0 (2):\penalty0 241--245, 2008.

\bibitem[Cortes and Vapnik(1995)]{cortes1995support}
Corinna Cortes and Vladimir Vapnik.
\newblock Support-vector networks.
\newblock \emph{Machine learning}, 20:\penalty0 273--297, 1995.

\bibitem[Amari and Wu(1999)]{amari1999improving}
Shun-ichi Amari and Si~Wu.
\newblock Improving support vector machine classifiers by modifying kernel functions.
\newblock \emph{Neural Networks}, 12\penalty0 (6):\penalty0 783--789, 1999.

\end{thebibliography}

\appendix

\section{The optimal linear binary classifier: SVM}
\label{appendixA}

We define the classification function as $f(x)=w^Tx+b$. Using Lagrange's multiplier method, the internal minimization problem of the variation problem in Equation (\ref{vsvmobj})  can be transformed into Equation (\ref{vsvmobjlag}).

\begin{equation}\label{vsvmobjlag}
J(f,x,x^{'},\lambda_T,\lambda_F)=\|x-x^{'}\|^2+\lambda_T(f(x)-1)+\lambda_F(f(x^{'})+1)
\end{equation}
In order to find the minimum value of $J$, it is necessary to obtain the derivative of $J$ on the unknown variables, that is, 

\begin{equation}\label{vsvmobjlagpx}
\begin{aligned}
\frac{\partial J}{\partial x}=2(x-x^{'})+\lambda_T\frac{\partial f}{\partial x}=0\\
\frac{\partial J}{\partial x^{'}}=2(x^{'}-x)+\lambda_F\frac{\partial f}{\partial x^{'}}=0
\end{aligned}
\end{equation}
By solving the system of equations in (\ref{vsvmobjlagpx}), the following results can be obtained
\begin{equation}\label{vsvmx}
\begin{aligned}
x^{'}=x+0.5\lambda_T\frac{\partial f}{\partial x}\\
x=x^{'}+0.5\lambda_F\frac{\partial f}{\partial x^{'}}
\end{aligned}
\end{equation}


Next, bring the results of Equation (\ref{vsvmx}) into the objective function of Equation (\ref{vsvmobj}), the following result can be obtained,
\begin{equation}\label{vsvmo}
 d(\Omega _-^f,\Omega _+^f)=\min_{f(x)=1,f(x^{'})=-1}\|x-x^{'}\|^2 =0.25\lambda_F^2\|\frac{\partial f}{\partial x^{'}}\|^2=0.25\lambda_T^2\|\frac{\partial f}{\partial x}\|^2, \forall k, f(x_k)y_k\geq 1.
\end{equation}

 It is known that $\frac{\partial f}{\partial x}=\frac{\partial f}{\partial x^{'}}=w$, then through Equation (\ref{vsvmobjlagpx}), the following results can be obtained
\begin{equation}\label{svmx}
\begin{aligned}
x^{'}=x+0.5\lambda_Tw\\
x=x^{'}+0.5\lambda_Fw
\end{aligned}
\end{equation}


By analyzing Equation (\ref{svmx}), we can get the relationship between $\lambda_T$ and $\lambda_F$, that is, $\lambda_T=-\lambda_F$. Considering that $f(x)=1$, then we can get the following results:

\begin{equation}\label{svm1}
f(x)=w^Tx + b = w^T(x^{'}+0.5\lambda_Fw)+b=w^Tx^{'}+0.5\lambda_Fw^Tw+b=f(x^{'})+0.5\lambda_Fw^Tw=1
\end{equation}
Considering that $f(x^{'})=-1$, we get the equation (\ref{svm2}).

\begin{equation}\label{svm2}
0.5\lambda_F=\frac{2}{w^Tw}=\frac{2}{\|w\|^2}
\end{equation}
By the Equation (\ref{vsvmo}) and (\ref{svm2}), we can get $d(\Omega _-^f,\Omega _+^f)=\frac{4}{\|w\|^4}\|w\|^2=\frac{4}{\|w\|^2} $. Therefore, the final optimization goal is as follows:

\begin{equation}\label{vsvmobjli}
f^*=\max_{f(x)=w^Tx+b} \min_{f(x)=1,f(x^{'})=-1} \|x-x^{'}\|^2=max_{w,b}\frac{4}{\|w\|^2}, \forall k, f(x_k)y_k\geq 1.
\end{equation}
It can be found that Equation (\ref{vsvmobjli}) is the optimization goal of SVM \citep{cortes1995support}. So, the variation problem defined by Equation (\ref{vsvmobj}) can deduce the objective function of the SVM algorithm for linear classification.

\section{Ellipsoid as classification boundary}
\label{ellipsoid surface}
 
Assume that the equation of the classification surface is expressed as $ f(\tilde{x})=\tilde{x}^T \Lambda \tilde{x}+\tilde{b}^T\tilde{x}+C$, where $\Lambda$ is a diagonal matrix, and the elements on the diagonal are all greater than 0. First, the surface equation can be transformed into the following equation:

\begin{eqnarray} \label{Ellipsoid}
\begin{aligned}
f(\tilde{x}) = \tilde{x}^T\Lambda\tilde{x}+\tilde{b}^T\tilde{x}+C & = \tilde{x}^T \Lambda \tilde{x}+\tilde{b}^T\Lambda^{-1/2}\Lambda^{1/2}\tilde{x}+C\\
& =  (\Lambda^{1/2}\tilde{x}+\frac{1}{2}\Lambda^{-1/2}\tilde{b})^T(\Lambda^{1/2}\tilde{x}+\frac{1}{2}\Lambda^{-1/2}\tilde{b})-\frac{1}{4} \tilde{b}^T\Lambda^{-1}\tilde{b}+C\\ 
& =  \left \|\Lambda^{1/2}\tilde{x}+\frac{1}{2}\Lambda^{-1/2}\tilde{b}   \right \|^2 -\frac{1}{4} \tilde{b}^T\Lambda^{-1}\tilde{b}+C\\ 
\end{aligned}
\end{eqnarray}

According to Equation 1, we need to obtain the support vectors of $\pm1$ classes respectively. The support vectors belonging to the positive class can be obtained by solving $f(\tilde{x})=1$, and result is as follows:

\begin{equation}
\begin{aligned}
&\left \|\Lambda ^{1/2}\tilde{x} +\frac{1}{2}\Lambda^{-1/2}\tilde{b}   \right \|^2 -\frac{1}{4} \tilde{b}^T\Lambda^{-1}\tilde{b}+C  = 1\\
&\left \|\Lambda^{1/2}\tilde{x}+\frac{1}{2}\Lambda^{-1/2}\tilde{b}   \right \|^2  = \frac{1}{4} \tilde{b}^T\Lambda^{-1}\tilde{b}-C+1\\
&\Lambda^{1/2}\tilde{x}+\frac{1}{2}\Lambda^{-1/2}\tilde{b}  = u\sqrt{\frac{1}{4} \tilde{b} ^T\Lambda^{-1}\tilde{b}-C+1}, \qquad u^Tu  = 1\\
&\tilde{x}  = -\frac{1}{2} \Lambda^{-1}\tilde{b}-\Lambda^{-1/2}u\sqrt{\frac{1}{4} \tilde{b}^T\Lambda^{-1}\tilde{b}-C+1}, \qquad u^Tu  = 1\\
\end{aligned}
\end{equation}

Similarly, the support vectors belonging to the negative class can be obtained by the following formula:

\begin{equation}
\begin{aligned}
&\left \|\Lambda ^{1/2}\tilde{x}^{'} +\frac{1}{2}\Lambda^{-1/2}\tilde{b}   \right \|^2 -\frac{1}{4} \tilde{b}^T\Lambda^{-1}\tilde{b}+C  = -1\\
&\left \|\Lambda^{1/2}\tilde{x}^{'}+\frac{1}{2}\Lambda^{-1/2}\tilde{b}   \right \|^2  = \frac{1}{4} \tilde{b}^T\Lambda^{-1}\tilde{b}-C-1\\
&\Lambda^{1/2}\tilde{x}^{'}+\frac{1}{2}\Lambda^{-1/2}\tilde{b}  = v\sqrt{\frac{1}{4} \tilde{b} ^T\Lambda^{-1}\tilde{b}-C-1}, \qquad v^Tv  = 1\\
&\tilde{x}^{'}  = -\frac{1}{2} \Lambda^{-1}\tilde{b}-\Lambda^{-1/2}v\sqrt{\frac{1}{4} \tilde{b}^T\Lambda^{-1}\tilde{b}-C-1}, \qquad v^Tv  = 1\\
\end{aligned}
\end{equation}

Let $\alpha =\sqrt{\frac{1}{4} \tilde{b}^T\Lambda^{-1}\tilde{b}-C+1}$ and $\beta =\sqrt{\frac{1}{4} \tilde{b}^T\Lambda^{-1}\tilde{b}-C-1}$, then according to our variation formula (\ref{vsvmobj}), we can get the following results:

\begin{align}
d(\Omega _-^f,\Omega _+^f)=\min_{f(\tilde{x})= 1,f(\tilde{x}^{'}=-1)} \left \| \tilde{x}-\tilde{x}^{'} \right \| ^2=
\min_{\left \| u \right \| =  1, \left \| v \right \|  =  1}
\left \| -\Lambda ^{-1/2}u\alpha +\Lambda ^{-1/2}v\beta \right \| ^2
\end{align}
With the help of the Lagrange's multiplier method, the following results can be obtained:

\begin{equation}
J(f,\tilde{x},\tilde{x}^{'},\lambda_T,\lambda_F)=\left \| -\Lambda ^{-1/2}u\alpha +\Lambda ^{-1/2}v\beta \right \| ^2
+\lambda_T(u^Tu-1)+\lambda_F(v^Tv-1)
\end{equation}
In order to find the minimum value of the above formula, we derive the derivatives of the variables respectively:

\begin{equation}
\begin{aligned}
\frac{\partial J}{\partial u}=-2\Lambda ^{-1/2}\alpha(-\Lambda^{-1/2}u\alpha +\Lambda^{-1/2}v\beta)+2\lambda _{T}u=0\\
\frac{\partial J}{\partial v}=2\Lambda^{-1/2}\beta (-\Lambda^{-1/2}u\alpha +\Lambda^{-1/2}v\beta)+2\lambda _{F}v=0
\end{aligned}
\end{equation}
Multiply the two equations in the above formula by $\beta$ and $\alpha$ respectively, and then add them together to get the following results:
\begin{eqnarray}
\lambda _{T}\beta u + \lambda _{F}\alpha v =  0\Rightarrow \lambda _{T}\beta u = -\lambda _{F}\alpha v
\end{eqnarray}
Since $u$ and $v$ are unit vectors, the solution to the above equation is:
\begin{eqnarray}
\left | \lambda _{T}\beta \right |  =\left | \lambda _{F}\alpha \right |,\quad u=v \quad or \quad u=-v
\end{eqnarray}
If $u=v$, then $J=(\alpha -\beta )^2u^{T}\Lambda^{-1}u$. If $u=-v$, then $J=(\alpha + \beta )^2u^{T}\Lambda^{-1}u$. Since $(\alpha + \beta )^2u^{T}\Lambda^{-1}u > (\alpha -\beta )^2u^{T}\Lambda^{-1}u$, the optimal solution of Equation (19) is $u=v$. So,

\begin{align}
d(\Omega _-^f,\Omega _+^f)=\min_{f(\tilde{x})= 1,f(\tilde{x}^{'}=-1)} \left \| \tilde{x}-\tilde{x}^{'} \right \| ^2=
\min_{\left \| u \right \| =  1} (\alpha -\beta )^2u^{T}\Lambda^{-1}u
\end{align}
We know that the minimum value of Equation (23) is $(\alpha -\beta )^2l_1^{-1}$ when $u=p_1$ ($p_1$ is the eigenvector corresponding to the maximum eigenvalue $l_1$ of matrix $A$). Therefore the final optimization goal is:
\begin{equation}
\begin{aligned}\label{e_loss1}
f^*=\max_{f(x)=x^TAx+b^Tx+C} \min_{f(x)=1,f(x^{'})=-1} \|x-x^{'}\|^2=max_{A,b,C}(\alpha -\beta )^2l_1^{-1} \\
 \forall k, f(x_k)y_k\geq 1
\end{aligned}
\end{equation}



\section{If the parabolic surface is used as the classification boundary, then $d(\Omega _-^f,\Omega _+^f) = 0$.}
\label{paraboloid surface}
Assuming that the classification boundary is $f(\tilde{x})= \tilde{x}_{[1:d]}^T \Lambda_{[1:d]} \tilde{x}_{[1:d]}+\tilde{b}^T\tilde{x}+C =0$. Similar to Equation (\ref{Ellipsoid}), we can transform the equation of the classification boundary into:

\begin{equation}\label{Paraboloid}
\begin{aligned}
f(\tilde{x}) & = \tilde{x}_{[1:d]}^T \Lambda_{[1:d]} \tilde{x}_{[1:d]}+\tilde{b}^T\tilde{x}+C\\
&=\left \|\Lambda_{[1:d]}^{1/2}\tilde{x}_{[1:d]}+\frac{1}{2}\Lambda_{[1:d]}^{-1/2}\tilde{b}_{[1:d]}   
\right \|^2 -\frac{1}{4} \tilde{b}_{[1:d]}^T\Lambda_{[1:d]}^{-1}\tilde{b}_{[1:d]}+
\tilde{b}_{[d+1:p]}^T\tilde{x}_{[d+1:p]}+C
\end{aligned}
\end{equation}

Based on previous experience, we first need to determine the support vectors. For the support vectors here, we assume we know $\tilde{x}_{[d+1:p]}$ and find $\tilde{x}_{[1:d]}$. The support vector belonging to each class is determined by the following formula:

\begin{eqnarray}
\begin{aligned}
\tilde{x}_{[1:d]} & = -\frac{1}{2} \Lambda_{[1:d]}^{-1}\tilde{b}_{[1:d]}-\Lambda_{[1:d]}^{-1/2}\gamma \sqrt{\frac{1}{4} \tilde{b}_{[1:d]}^T\Lambda_{[1:d]}^{-1}\tilde{b}_{[1:d]}-\tilde{b}_{[d+1:p]}^T\tilde{x}_{[d+1:p]}-C+1}, \quad \gamma^T\gamma   = 1\\
\tilde{x}_{[1:d]}^{'} & = -\frac{1}{2} \Lambda_{[1:d]}^{-1}\tilde{b}_{[1:d]}-\Lambda_{[1:d]}^{-1/2}\eta \sqrt{\frac{1}{4} \tilde{b}_{[1:d]}^T\Lambda_{[1:d]}^{-1}\tilde{b}_{[1:d]}-\tilde{b}_{[d+1:p]}^T\tilde{x}_{[d+1:p]}^{'}-C-1}, \quad \eta ^T\eta  = 1 
\end{aligned}
\end{eqnarray}

Suppose $\tilde{x}_{[d+1:p]}=\tilde{x}_{[d+1:p]}^{'}$, then the distance between these two support vectors is as follows:

\begin{eqnarray}\label{dist_paowu}
\begin{aligned}
 \|x-\hat{x} \|^2&=\|\tilde{x}_{[1:d]}-\tilde{x}_{[1:d]}^{'}\|^2=\left \| -\Lambda_{[1:d]} ^{-1/2}\gamma a+\Lambda_{[1:d]} ^{-1/2}\eta b\right \| ^2\\
a&=\sqrt{\frac{1}{4} \tilde{b}_{[1:d]}^T\Lambda_{[1:d]}^{-1}\tilde{b}_{[1:d]}-\tilde{b}_{[d+1:p]}^T\tilde{x}_{[d+1:p]}-C+1}\\
b&=\sqrt{\frac{1}{4} \tilde{b}_{[1:d]}^T\Lambda_{[1:d]}^{-1}\tilde{b}_{[1:d]}-\tilde{b}_{[d+1:p]}^T\tilde{x}_{[d+1:p]}^{'}-C-1}\\
\end{aligned}
\end{eqnarray}

With the help of the derivation process in the elliptical quadratic surface, we know that the minimum value of Equation (\ref{dist_paowu}) is $(a-b)^2\gamma ^{T}\Lambda_{[1:d]}^{-1}\gamma$. Since the support vector we consider here is a special type of support vector, the minimum value $(a-b)^2\gamma ^{T}\Lambda_{[1:d]}^{-1}\gamma$ is an upper bound of $d(\Omega _-^f,\Omega _+^f)$, that is, $d(\Omega _-^f,\Omega _+^f)\le (a-b)^2\gamma ^{T}\Lambda_{[1:d]}^{-1}\gamma$. However, $a$ and $b$ are related to $x$ and cannot be considered a constant. When $\tilde{x}_{[d+1:p]}(=\tilde{x}_{[d+1:p]}^{'})$ approaches infinity, the limit of $a-b$ is 

\begin{eqnarray}
\begin{aligned}
\lim_{x_{[d+1:p]} \to \infty } a-b&=\lim_{x_{[d+1:p]} \to \infty }\frac{2}{a+b} =0
\end{aligned}
\end{eqnarray}
Therefore, $d(\Omega _-^f,\Omega _+^f) = \min_{f(\tilde{x})= 1,f(\tilde{x}^{'}=-1)} \left \| \tilde{x}-\tilde{x}^{'} \right \| ^2=0$. If so, the variational problem proposed in this article will not be optimized, that is, we cannot complete the binary classification task. This result shows that parabolic surfaces cannot be used for classification, which is contradictory to the actual situation. Therefore, the variational definition based on $2$-norm cannot be generalized to parabolic surfaces.


\section{If the hyperbolic surface is used as the classification boundary, then $d(\Omega _-^f,\Omega _+^f) = 0$}
\label{hyperbolic surface}
Assume that the equation of the classification surface is expressed as $ f(\tilde{x})=\tilde{x}^T \Lambda \tilde{x}+\tilde{b}^T\tilde{x}+C$, where $\Lambda$ is a diagonal matrix, and the elements on the diagonal are $l_1\ge l_2\ge ...l_d > 0 > l_{d+1} \ge ... \ge l_{p}$. We can transform the equation of the classification boundary into:
\begin{eqnarray} \label{hyperbolic}
\begin{aligned}
\tilde{x}^T\Lambda\tilde{x}+\tilde{b}^T\tilde{x}+C & = \begin{bmatrix}
 \tilde{x}_{[1:d]} &\tilde{x}_{[d+1:p]}
\end{bmatrix}\begin{bmatrix}
 \Lambda_1  &0 \\
  0&\Lambda_2 
\end{bmatrix}\begin{bmatrix}
 \tilde{x}_{[1:d]}\\
\tilde{x}_{[d+1:p]}
\end{bmatrix} + \tilde{b}^T\begin{bmatrix}
 \tilde{x}_{[1:d]}\\
\tilde{x}_{[d+1:p]}
\end{bmatrix}+C\\
&=\tilde{x}_{[1:d]}^T\Lambda_1\tilde{x}_{[1:d]}+\tilde{b}_{[1:d]}^T\tilde{x}_{[1:d]} + \tilde{x}_{[d+1:p]}^T\Lambda_2\tilde{x}_{[d+1:p]}+\tilde{b}_{[d+1:p]}^T\tilde{x}_{[d+1:p]}+C\\
& =  \left \|\Lambda_1^{1/2}\tilde{x}_{[1:d]}+\frac{1}{2}\Lambda_1^{-1/2}\tilde{b}_{[1:d]}   \right \|^2 -\frac{1}{4} \tilde{b}_{[1:d]}^T\Lambda_1^{-1}\tilde{b}_{[1:d]}+ \tilde{x}_{[d+1:p]}^T\Lambda_2\tilde{x}_{[d+1:p]}\\
&\qquad +\tilde{b}_{[d+1:p]}^T\tilde{x}_{[d+1:p]}+C\\ 
\end{aligned}
\end{eqnarray}
where $\Lambda_1=diag([l_1, l_2, ...,l_d])$, and $\Lambda_2=diag([l_{d+1}, l_{d+2}, ...,l_p])$.

Similar to the proof process in Appendix \ref{paraboloid surface}, here we also look for specific support vectors (assuming that $\tilde{x}_{[d+1:p]}$ is already known, find the remaining coordinate values $\tilde{x}_{[1:d]}$). The support vector belonging to each class is determined by the following formula:

\begin{eqnarray}
\begin{aligned}
\tilde{x}_{[1:d]} & = -\frac{1}{2} \Lambda_1^{-1}\tilde{b}_{[1:d]}-\Lambda_1^{-1/2}\gamma \sqrt{\frac{1}{4} \tilde{b}_{[1:d]}^T\Lambda_1^{-1}\tilde{b}_{[1:d]}-D-C+1}, \quad \gamma^T\gamma   = 1\\
\tilde{x}_{[1:d]}^{'} & = -\frac{1}{2} \Lambda_1^{-1}\tilde{b}_{[1:d]}-\Lambda_1^{-1/2}\eta \sqrt{\frac{1}{4} \tilde{b}_{[1:d]}^T\Lambda_1^{-1}\tilde{b}_{[1:d]}-D^{'}-C-1}, \quad \eta ^T\eta  = 1 
\end{aligned}
\end{eqnarray}
where $D=\tilde{x}_{[d+1:p]}^T\Lambda_2\tilde{x}_{[d+1:p]}+\tilde{b}_{[d+1:p]}^T\tilde{x}_{[d+1:p]}$, and $D^{'}=\tilde{x}_{[d+1:p]}^{'T}\Lambda_2\tilde{x}_{[d+1:p]}^{'}+\tilde{b}_{[d+1:p]}^T\tilde{x}_{[d+1:p]}^{'}$. Suppose $\tilde{x}_{[d+1:p]}=\tilde{x}_{[d+1:p]}^{'}$, then $D=D^{'}$ and the distance between the support vectors is determined by the following formula

\begin{eqnarray}\label{dist_hyperbolic}
\begin{aligned}
 \|x-\hat{x} \|^2&=\|\tilde{x}_{[1:d]}-\tilde{x}_{[1:d]}^{'}\|^2=\left \| -\Lambda_1 ^{-1/2}\gamma a+\Lambda_1 ^{-1/2}\eta b\right \| ^2\\
a&=\sqrt{\frac{1}{4} \tilde{b}_{[1:d]}^T\Lambda_1^{-1}\tilde{b}_{[1:d]} - D-C+1}\\
b&=\sqrt{\frac{1}{4} \tilde{b}_{[1:d]}^T\Lambda_1^{-1}\tilde{b}_{[1:d]} - D-C-1}\\
\end{aligned}
\end{eqnarray}
There is no essential difference between the above formula and Equation (\ref{dist_paowu}) in Appendix \ref{paraboloid surface}. Therefore, the minimum value of Equation (\ref{dist_hyperbolic}) is $(a-b)^2\gamma ^{T}\Lambda_1^{-1}\gamma$. Since the support vector we consider here is a special type of support vector, the minimum value $(a-b)^2\gamma ^{T}\Lambda_1^{-1}\gamma$ is an upper bound of $d(\Omega _-^f,\Omega _+^f)$, that is, $d(\Omega _-^f,\Omega _+^f)\le (a-b)^2\gamma ^{T}\Lambda_1^{-1}\gamma$. Since the diagonal elements of $\Lambda_2$ are all less than zero, $-D$ tends to positive infinity when $\tilde{x}_{[d+1:p]}$ tends to infinity. We know that $a$ and $b$ are related to $D$, so

\begin{eqnarray}
\begin{aligned}
\lim_{x_{[d+1:p]} \to \infty } a-b&=\lim_{x_{[d+1:p]} \to \infty }\frac{2}{a+b} =0
\end{aligned}
\end{eqnarray}
Therefore, $d(\Omega _-^f,\Omega _+^f) = \min_{f(\tilde{x})= 1,f(\tilde{x}^{'}=-1)} \left \| \tilde{x}-\tilde{x}^{'} \right \| ^2=0$. 
\end{document}